\documentclass[11pt]{article}

\usepackage{graphicx}
\usepackage{xcolor}
\usepackage{geometry}
\usepackage{amsmath}
\usepackage{amssymb}
\usepackage{booktabs}
\usepackage{multirow}
\usepackage{array}
\usepackage{enumitem}
\usepackage[protrusion=true,expansion=false]{microtype}
\usepackage{xurl}
\usepackage{float}
\usepackage{caption}
\usepackage{tcolorbox}

\usepackage[numbers,super,sort&compress]{natbib}
\usepackage[
    colorlinks=true,
    linkcolor=blue,
    citecolor=darkgray,
    urlcolor=blue,
    bookmarks=true,
    bookmarksnumbered=true,
    unicode=true
]{hyperref}
\hypersetup{
    pdftitle={From BERT to Frontier Agents: Eight Years of Language-Model Progress},
    pdfauthor={Pranav Kumar Kaliaperumal <prka5235@colorado.edu>},
    pdfsubject={cs.CL, cs.AI, cs.LG},
    pdfkeywords={large language models, benchmarks, scaling, cost efficiency, task specialization, self-consistency, model routing}
}

\setlist[itemize]{itemsep=0.2em, leftmargin=1.4em}
\setlist[enumerate]{itemsep=0.2em, leftmargin=1.6em}

\newcommand{\submissiongraphic}[2][]{%
  \IfFileExists{figures/#2}{%
    \includegraphics[#1]{figures/#2}%
  }{%
    \IfFileExists{#2}{%
      \includegraphics[#1]{#2}%
    }{%
      \PackageWarningNoLine{llm-progression}{Missing figure #2; using a placeholder}%
      \fbox{\begin{minipage}[c][1.6in][c]{0.88\linewidth}
        \centering Figure file \texttt{#2} was not included in the submission.
      \end{minipage}}%
    }%
  }%
}

\title{\textbf{From BERT to Frontier Agents: Eight Years of Language-Model\\Progress, the Collapse of the Capability--Cost Curve,\\and the Rise of Task-Targeted Models}\\[0.6em]
\large An empirical review with reproducible measurements, 2018--2026}

\author{Pranav Kumar Kaliaperumal\\
\texttt{prka5235@colorado.edu}}
\date{August 2026}

\begin{document}
\maketitle

\begin{abstract}
\noindent
Between October 2018 and July 2026, language models moved from BERT, a
340-million-parameter encoder that could barely generate text, to
trillion-parameter agentic systems that solve International Mathematical
Olympiad problems, repair software repositories, and operate desktop
computers. We trace that progression and measure its practical consequences
with public benchmarks, prices, and reproducible experiments. A log-linear fit
to SWE-bench Verified estimates ${\sim}5.8\times$ annual growth in the odds
of resolving a real GitHub issue since October 2024. Over the same period, the
capability--cost curve collapsed: GPT-5.6 Luna, OpenAI's budget tier at
\$1/\$6 per million input/output tokens, matches or beats the eleven-week-old
GPT-5.5 flagship on most agentic and professional-work benchmarks in the
vendor's tables at one-fifth the token price. It also exceeds the August 2025
GPT-5 SWE-bench Verified score by ${\sim}18$ points in our reconstruction of
an independent harness. Task-level results reveal a fragmented frontier:
Claude Opus~5 leads human-preference frontend coding, Claude Fable~5 leads
repository-level coding, GPT-5.6 Sol leads agentic terminal work, and Opus~5
quadrupled the ARC-AGI-3 record. A two-model router recovers the full
per-benchmark gain of a six-model oracle on our comparison suite. We also run
a traced grade-school-math experiment. After selecting the prompt, temperature,
model size, and selector on a locked 16-item development split, we freeze a
Qwen2.5-1.5B configuration and evaluate it once on 100 disjoint items. Greedy
decoding solves 58/100; four-sample plurality and a prespecified verifier each
solve 62/100 (paired exact McNemar $p=0.481$), while the any-sample oracle
reaches 79/100. A post-evaluation confidence model ranks the vote-at-four
outputs with five-fold out-of-fold ROC AUC 0.833 and a Brier score of 0.147,
versus 0.236 for a prevalence baseline. Its highest-confidence 50 predictions
contain 47 correct answers, suggesting a useful triage signal rather than a
new full-coverage accuracy gain. The repository includes all data, code,
complete traces, tests, and compilation scripts.
\end{abstract}

\vspace{0.5em}
\noindent\textbf{Keywords:} large language models; evaluation; scaling;
inference economics; task specialization; self-consistency; confidence estimation; model routing

\tableofcontents
\newpage

\section{Introduction}\label{sec:intro}

In October 2018, Google released BERT, a bidirectional Transformer encoder that
set the state of the art on eleven natural-language understanding tasks and
pushed the GLUE benchmark to 80.5\%~\cite{devlin2019bert}. BERT could not write
a coherent paragraph; it was a \emph{reading} model, fine-tuned per task. In
July 2026, Anthropic released Claude Opus~5 and reported 42/42 under its
post-contest, no-tools grading protocol for the 2026 International Mathematical
Olympiad problems; ARC Prize independently measured a roughly fourfold jump on
ARC-AGI-3~\cite{anthropic2026opus5card,arcprize2026opus5}. Between those
two releases sit eight years in which the field's dominant paradigm changed at
least four times: from task-specific fine-tuning, to few-shot prompting of
large pretrained decoders, to instruction tuning and human-feedback alignment,
to reasoning models and tool-using agents.

We ask three related questions. How did the field move from BERT to the
2026 frontier of GPT-5.6, Claude Fable~5 and Opus~5, Kimi~K3, and their
contemporaries? How quickly did capability and cost efficiency improve when
measured from public benchmark and pricing series rather than impressions?
And how much more performance can a practitioner obtain from fixed model
weights by spending inference compute carefully, calibrating reasoning effort,
estimating confidence, and routing work across models?

\paragraph{Contributions.}
\begin{enumerate}
    \item A consolidated, sourced timeline of language-model capability from
    2018 to 2026 across three benchmark eras (Section~\ref{sec:history},
    Table~\ref{tab:timeline}).
    \item Trend fits showing ${\sim}5.8\times$ annual growth in the odds of
    solving SWE-bench Verified tasks, alongside the saturation of
    knowledge benchmarks such as MMLU (Section~\ref{sec:capability}).
    \item A cost analysis showing a ${\sim}60\times$ decline in input-token
    prices from GPT-3 (2020) to GPT-5.6 Luna (2026), and a demonstration that
    today's budget tier matches the flagship of one quarter ago on most
    agentic/professional benchmarks (Section~\ref{sec:cost}).
    \item Evidence that the 2026 frontier is task-fragmented, and that simple
    model routing recovers the per-task optimum
    (Section~\ref{sec:specialization}).
    \item A locked development/evaluation study of self-consistency for a
    small open model, followed by an exploratory confidence model that ranks
    predictions for triage. Complete traces, paired uncertainty, and
    reasoning-effort analyses accompany the results
    (Section~\ref{sec:improvements}).
\end{enumerate}

\section{Eight years in four paradigms}\label{sec:history}

\subsection{The encoder era: BERT and the fine-tuning paradigm (2018--2019)}
BERT~\cite{devlin2019bert} established the pretrain-then-fine-tune recipe:
masked-language-model pretraining on unlabeled text, followed by a small
task-specific head. RoBERTa~\cite{liu2019roberta} showed the recipe was
undertrained rather than wrong, reaching 88.5 on GLUE with more data and longer
training; T5~\cite{raffel2020t5} unified all tasks as text-to-text and reached
90.3 on SuperGLUE. GPT-2~\cite{radford2019gpt2} demonstrated that a
left-to-right decoder trained on enough web text acquired tasks
\emph{without} fine-tuning, but its zero-shot quality was not yet competitive.
The era's signature: capability lived in task-specific heads, and benchmarks
(GLUE, SuperGLUE, SQuAD) saturated within roughly two years of introduction.

\subsection{Scale and in-context learning: GPT-3 (2020--2021)}
GPT-3~\cite{brown2020gpt3} scaled the decoder recipe to 175B parameters and
showed that few-shot prompting --- a handful of examples in the context window
--- could substitute for gradient updates on many tasks. Its 43.9\%
five-shot MMLU~\cite{hendrycks2021mmlu} looks modest today, but the conceptual
shift was permanent: capability could be \emph{elicited} from a frozen general
model. The limits were equally clear: GPT-3 few-shot scored 71.8 on SuperGLUE,
well below fine-tuned systems, and its unaligned samples were unreliable.

\subsection{Alignment and instruction following (2022--2023)}
InstructGPT~\cite{ouyang2022instructgpt} applied reinforcement learning from
human feedback (RLHF) to make models follow instructions; ChatGPT (November
2022) packaged this for general users. GPT-4 (March 2023) jumped to 86.4\% on
MMLU~\cite{openai2023gpt4}, near the estimated human-expert level, and
established the modern pattern of a capability report accompanying release.
Meta's Llama series opened weights and created an open ecosystem; Claude and
Gemini established a multi-vendor frontier.

\subsection{Reasoning and agents (2024--2026)}
Test-time compute changed the scaling target. OpenAI's o1 (December 2024)
showed that longer, trained reasoning traces and additional inference tokens
could lift math and coding performance; its 91.8\% MMLU score effectively
saturated that benchmark~\cite{benchlm_mmlu}. Agency then changed the task
itself. SWE-bench Verified~\cite{jimenez2024swebench,openai2024swebenchverified}
made ``resolve a real GitHub issue'' the standard hard problem. Scores on
independent harnesses climbed from 49\% for Claude 3.5 Sonnet in October 2024
to 97\% for Claude Opus~5 in July 2026~\cite{vals2026swebench}. By mid-2026
the flagship
releases --- Claude Fable~5 (July 1), GPT-5.6 Sol/Terra/Luna (July 9), Kimi~K3
(July 16), Claude Opus~5 (July 24) --- were evaluated primarily on agentic,
long-horizon, and economically-weighted tasks rather than static question
answering~\cite{anthropic2026fable5,openai2026gpt56,moonshot2026k3,anthropic2026opus5}.
Kimi~K3 additionally marked the near-closure of the open--closed gap: a
2.8-trillion-parameter open-weight mixture-of-experts model competitive with
the best proprietary systems~\cite{moonshot2026k3}.

Table~\ref{tab:timeline} summarizes the arc, and Figure~\ref{fig:swe} shows
the agentic benchmark series at its current endpoint.

\begin{table}[t]
\centering
\caption{Selected milestones, 2018--2026. Scores are as originally reported by
the cited source; benchmark variants are noted.}\label{tab:timeline}
\small
\begin{tabular}{@{}llllr@{}}
\toprule
Date & Model & Org & Milestone metric & Score \\
\midrule
2018-10 & BERT-Large & Google & GLUE & 80.5 \\
2019-07 & RoBERTa & Meta & GLUE & 88.5 \\
2019-10 & T5-11B & Google & SuperGLUE & 90.3 \\
2020-05 & GPT-3 & OpenAI & MMLU (5-shot) & 43.9 \\
2023-03 & GPT-4 & OpenAI & MMLU (5-shot) & 86.4 \\
2024-05 & GPT-4o & OpenAI & MMLU & 88.7 \\
2024-12 & o1 & OpenAI & MMLU & 91.8 \\
2024-10 & Claude 3.5 Sonnet & Anthropic & SWE-bench Ver. & 49.0 \\
2025-05 & Claude Opus 4 & Anthropic & SWE-bench Ver. & 72.5 \\
2025-08 & GPT-5 & OpenAI & SWE-bench Ver. & 74.9 \\
2025-11 & Claude Opus 4.5 & Anthropic & SWE-bench Ver. & 76.8 \\
2026-04 & GPT-5.5 & OpenAI & SWE-bench Ver. (vendor) & 85.1 \\
2026-07 & Claude Fable 5 & Anthropic & SWE-bench Ver. (indep.) & 95.0 \\
2026-07 & GPT-5.6 Sol & OpenAI & SWE-bench Ver. (indep.) & 96.2 \\
2026-07 & Kimi K3 & Moonshot & SWE-bench Ver. (indep.) & 93.4 \\
2026-07 & Claude Opus 5 & Anthropic & SWE-bench Ver. (indep.) & 97.0 \\
\bottomrule
\end{tabular}
\end{table}

\section{Data and methods}\label{sec:methods}

\subsection{Data}
The repository contains three analysis tables, a locked evaluation split, and
the complete traces needed to reproduce every reported result.

\paragraph{D1: Timeline.} Twenty-seven (model, benchmark, score, date, source)
rows spanning 2018--2026, drawn from the original papers and from leaderboard
mirrors (Table~\ref{tab:timeline}; full file \texttt{data/models\_timeline.csv}).
Where multiple evaluation variants exist we record the variant explicitly
(e.g., vendor-reported vs.\ independent SWE-bench Verified).

\paragraph{D2: 2026 frontier matrix.} Per-model benchmark scores and API prices
for the July 2026 frontier --- GPT-5.6 Sol/Terra/Luna, Claude Fable~5, Claude
Opus~5, Kimi~K3, plus the April 2026 flagship GPT-5.5 and Claude Opus~4.8 ---
from OpenAI's GPT-5.6 launch tables~\cite{openai2026gpt56}, Anthropic's Opus~5
launch and system card~\cite{anthropic2026opus5,anthropic2026opus5card},
Artificial Analysis~\cite{aa2026gpt56,gist2026aa56}, the Vals AI independent
SWE-bench Verified harness~\cite{vals2026swebench}, and arena.ai's
human-preference Frontend Code Arena~\cite{arena2026webdev}.

\paragraph{D3: Pricing.} Input/output prices per million tokens for
representative API models from 2020 to 2026, using vendor documentation where
available~\cite{deploybase2026prices,openai2026gpt56}.

\paragraph{D4: Live-evaluation items.} The committed official GSM8K test split
contains 1,319 question/answer pairs~\cite{cobbe2021gsm8k}. We retain the
original 16 pilot identifiers as a locked development split and sample 100
evaluation identifiers once from the remaining rows
(\texttt{random\_state=20260810}) before v1 tuning. The split manifest records
both identifier lists and the source-parquet SHA-256 digest.

\paragraph{Provenance caveat.} Mid-2026 numbers mix vendor-reported and
independent evaluations. We label provenance wherever it matters and prefer
third-party measurements (Vals AI, Artificial Analysis, ARC Prize) when both
exist. Vendor launch numbers are claims, not independent results; we treat them
as such and note where vendor and third-party harnesses differ.

\begin{figure}[t]
\centering
\submissiongraphic[width=\linewidth]{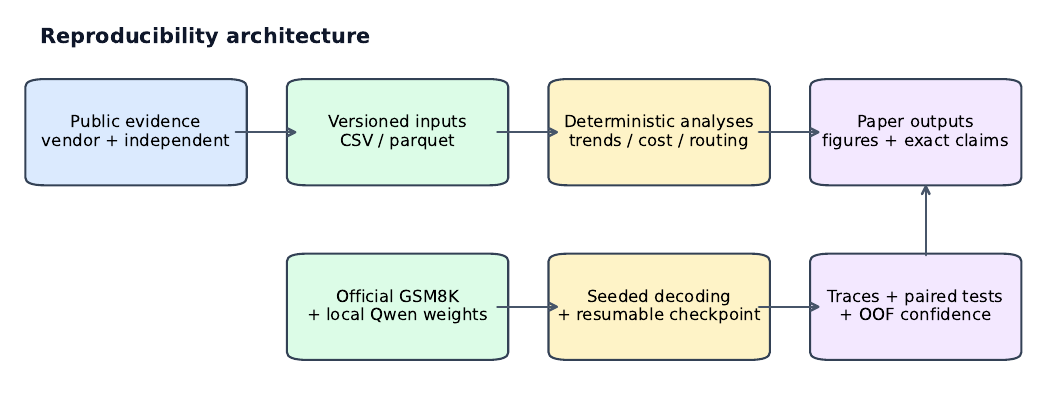}
\caption{Reproducibility architecture. Public evidence feeds deterministic
trend, cost, and routing analyses. The live evaluation records complete traces
and paired statistics; the exploratory confidence stage consumes those
versioned traces. Every path produces auditable paper outputs.}\label{fig:architecture}
\end{figure}

\subsection{Analysis methods}
\paragraph{Trend fitting.} For the SWE-bench Verified series we fit a
log-linear model to the \emph{odds} of solving a task,
$\mathrm{logit}(p)$, against release date, and report the implied annual
multiplicative growth of those odds with the fit $R^2$
(Figure~\ref{fig:swe}). For MMLU we compare the raw series with the estimated
human-expert level of 89.8\%~\cite{hendrycks2021mmlu}.

\paragraph{Pareto analysis.} Using Artificial Analysis's all-effort dataset for
the GPT-5.6 family (15 model$\times$effort settings with intelligence-index
scores and per-task costs~\cite{gist2026aa56,aa2026gpt56}), we compute the
cost--capability Pareto frontier and identify dominated settings.

\paragraph{Routing oracle.} On the fourteen benchmarks that OpenAI's GPT-5.6
launch table reports for all six of Sol, Terra, Luna, GPT-5.5, Fable~5, and
Opus~4.8, we normalize each benchmark to its best model (100), then compare
(i) the best single model's mean, (ii) an oracle that picks the best model per
benchmark, and (iii) restricted routers (a cheapest-within-one-point router
and a two-model router).

\paragraph{Live experiment.} Section~\ref{sec:selfcons} describes a controlled
development/evaluation protocol. On development data we screen three prompts
(official chat-template concise, four-shot, and solve-and-check) at
temperatures 0.4 and 0.7 using Qwen2.5-0.5B-Instruct, confirm the leading
four-shot/0.4 setting on all 16 items, and compare it with
Qwen2.5-1.5B-Instruct. The primary selector is four-sample plurality; a greedy
candidate verifier is prespecified as secondary. Before evaluation, a manifest
freezes Qwen2.5-1.5B-Instruct, the official chat template with four worked
examples, $k=4$, temperature 0.4, top-$p=0.95$, seed 1234, and the plurality
selector. We then run that configuration once on all 100 disjoint evaluation
items. Every record stores the greedy completion, four sampled completions,
canonical numeric answers, prefix votes, verifier decision, and correctness.
We report exact counts, Wilson 95\% intervals, the any-sample oracle ceiling,
and two-sided exact McNemar tests against paired greedy outcomes. Atomic
checkpoints, JSONL traces, split/config hashes, and integrity tests are
provided.
\paragraph{Exploratory confidence model.} After the frozen evaluation labels
were available, we specified a simple model to estimate whether vote-at-four
was correct. Its seven reference-free features are plurality share, normalized
answer entropy, the number of distinct sampled answers, agreement of plurality
with greedy and verifier outputs, and the mean and standard deviation of
sample-completion length. We standardize these features inside each training
fold and fit an L2-regularized logistic regression. Five-fold stratified
cross-validation with shuffling and seed 20260811 yields one out-of-fold
probability for every item. We report ROC AUC, Brier score against a constant
0.62 prevalence baseline, and accuracy among the predictions retained at fixed
50\% and 75\% coverage. The features, folds, metrics, and coverage levels were
set before scoring this added analysis, but the analysis itself was designed
after the v1 outcomes were known. It is exploratory and does not constitute a
second held-out evaluation.

\section{Results}\label{sec:results}

\subsection{Capability growth: three benchmark eras}\label{sec:capability}

\paragraph{Knowledge benchmarks saturated.} The MMLU series rose from 43.9\%
(GPT-3, 2020) to 86.4\% (GPT-4, 2023) to 91.8\% (o1, late
2024), past the 89.8\% estimated human-expert level. By 2024 the benchmark had
lost discriminative power: four strong models clustered within 4
points~\cite{wang2024mmlupro}, and benchmark mirrors now mark MMLU as
``saturated'' and exclude it from scoring~\cite{benchlm_mmlu}. The field
responded with harder knowledge tests (MMLU-Pro~\cite{wang2024mmlupro},
Humanity's Last Exam) and, increasingly, with agentic tasks.

\begin{figure}[t]
\centering
\submissiongraphic[width=0.9\linewidth]{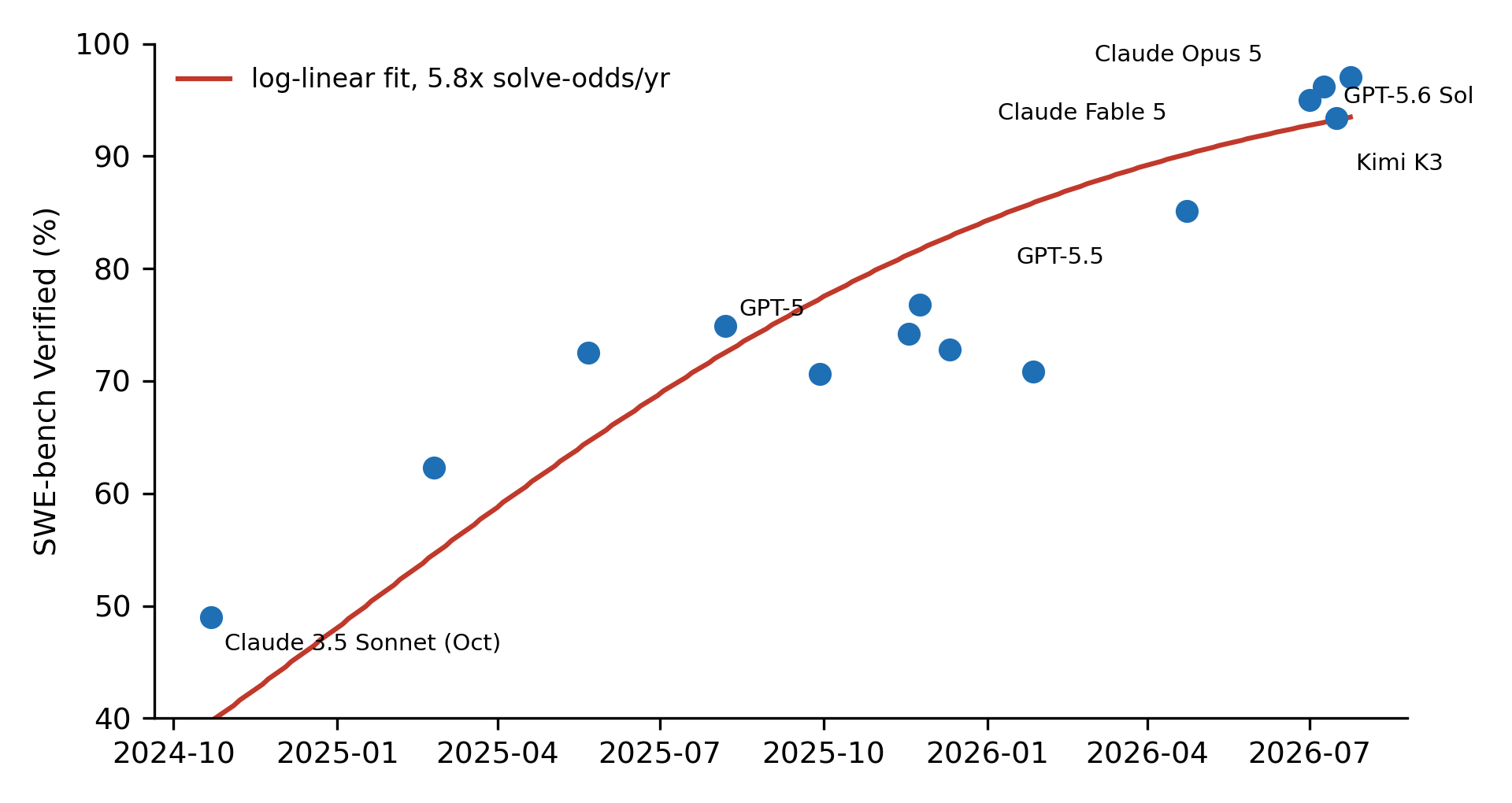}
\caption{SWE-bench Verified, October 2024 -- July 2026 (mix of vendor-reported
and independent harness results; see Table~\ref{tab:timeline}). The log-linear
fit on solve odds implies ${\sim}5.8\times$ growth per year.}\label{fig:swe}
\end{figure}

\paragraph{Agentic coding became the frontier metric --- and moved fast.}
Figure~\ref{fig:swe} shows SWE-bench Verified from October 2024 to July 2026.
A log-linear fit on the solve odds gives
$\mathrm{odds}_{t+1\mathrm{yr}} \approx 5.8\times\,\mathrm{odds}_t$
($R^2 = 0.78$, $n = 14$): the odds of resolving a real GitHub issue multiplied
by almost six each year. Independent evaluation puts the current frontier at
96--97\%~\cite{vals2026swebench}, implying the benchmark itself will saturate
within roughly a year --- the same lifecycle GLUE and MMLU went through, on a
harder task. Harder variants (SWE-bench Pro, Frontier-Bench, on which even the
best model scores only 43--64\%~\cite{anthropic2026opus5,openai2026gpt56}) are
already taking over.

\paragraph{Benchmark turnover accelerated.} GLUE remained discriminative
for about two years and MMLU for roughly four; SWE-bench Verified appears
headed toward a two-year lifespan. Capability keeps advancing, but each
measuring stick loses headroom soon after laboratories optimize against it.

\subsection{The collapse of the capability--cost curve}\label{sec:cost}

API input prices fell from GPT-3's \$60 per million tokens (2020) to GPT-5.6
Luna's \$1 (2026): a ${\sim}60\times$ decline,
compounding at roughly $2\times$ per year, \emph{while} capability itself
was compounding at a similar rate. The two effects multiply: a dollar buys
orders of magnitude more capability than a dollar in 2023.

\paragraph{Budget tier vs.\ last quarter's flagship.} The sharpest illustration
is internal to OpenAI's own July 2026 launch table~\cite{openai2026gpt56}.
GPT-5.6 Luna (\$1/\$6 per M input/output tokens at the official standard API
rate~\cite{openai2026gpt56}) against GPT-5.5 (\$5/\$30), the flagship released
eleven weeks earlier: across the 23 benchmarks reported for both, Luna wins 9
and loses 14 --- but the wins concentrate exactly where the economic value is.
On agentic and professional work (Agents' Last Exam, GDPval-AA, SWE-bench Pro,
DeepSWE, HealthBench Professional, AutomationBench, GraphWalks) Luna wins 7 of
10; it trails mainly on academic knowledge (GPQA), competition math
(FrontierMath), multimodal understanding, and --- sharply --- long-context
recall (41.3\% vs.\ 81.5\% on MRCR). Against the flagship of \emph{one year}
ago the comparison is not close: our reconstruction of overall SWE-bench
Verified from Vals AI's per-difficulty rows gives Luna 92.9\% vs.\ GPT-5's
74.9\% vendor-reported score in August
2025~\cite{vals2026swebench,localaimaster2026gpt55}. In practical terms,
\textbf{capability that was flagship-class 2--12 months ago now costs
${\sim}20\%$ of the flagship token price}. The cheapest tier still gives up
long-context recall and the hardest academic reasoning.

\paragraph{The mid tier is dead.} Our Pareto analysis of the fifteen
GPT-5.6 settings reproduces and confirms Artificial Analysis's
finding~\cite{gist2026aa56,aa2026gpt56}: every Terra configuration is dominated
--- there is always a Luna or Sol setting that is smarter at no extra cost or
equally smart and cheaper. Six of fifteen settings lie off the frontier. The
economic structure of a model family is now bimodal: a volume tier and a
premium tier, with the middle squeezed out.

\subsection{The frontier fragmented by task}\label{sec:specialization}

\begin{figure}[t]
\centering
\submissiongraphic[width=0.95\linewidth]{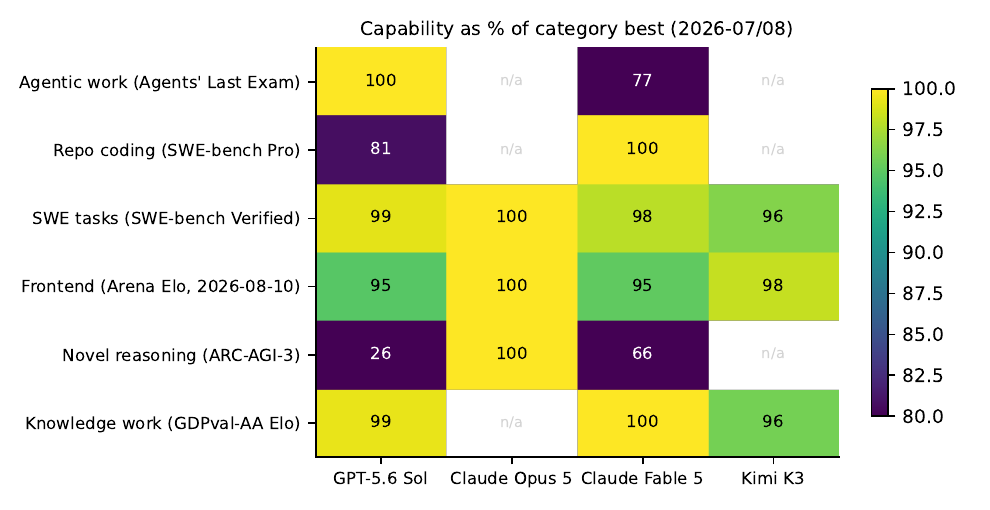}
\caption{Capability as a percentage of the category best, July--August 2026.
No column is uniformly dark: each frontier model has at least one category
where another model clearly leads. Cells marked n/a lack a comparable public
number --- itself a symptom of vendors reporting different
benchmarks.}\label{fig:heatmap}
\end{figure}

No single model leads the 2026 frontier; leadership is task-dependent
(Figure~\ref{fig:heatmap}):

\begin{itemize}
    \item \textbf{Frontend and design-oriented coding: Claude Opus 5.} On the
    live arena.ai WebDev leaderboard checked August 10, Opus~5 led blind
    developer preference at 1,712 Elo, with open-weight Kimi K3 second at
    1,682; both values were preliminary and will move as votes
    accumulate~\cite{arena2026webdev}.
    \item \textbf{Repository-level coding: Claude Fable 5.} Fable~5 leads
    SWE-bench Pro at 80.0\%, 15.4 points ahead of GPT-5.6 Sol
    (64.6\%)~\cite{openai2026gpt56}.
    \item \textbf{Agentic terminal and long-horizon professional work:
    GPT-5.6 Sol.} Sol leads the Artificial Analysis Coding Agent Index (80),
    Terminal-Bench 2.1 (88.8\%; 91.9\% in multi-agent Ultra mode), and Agents'
    Last Exam (52.7\%, +12.2 over the best non-OpenAI
    model)~\cite{aa2026gpt56,openai2026gpt56}.
    \item \textbf{Novel reasoning and computer use: Claude Opus 5.} Opus~5
    scored 30.2\% on ARC-AGI-3 --- nearly $4\times$ the previous record of
    7.8\% (GPT-5.6 Sol) --- plus a vendor-reported 42/42 on IMO 2026 under
    Anthropic's grading protocol and the lead on
    OSWorld~2.0 computer use~\cite{anthropic2026opus5,anthropic2026opus5card,arcprize2026opus5}.
    \item \textbf{Best-value agentic work: Claude Opus 5 again,} reaching
    within 0.5\% of Fable~5's CursorBench peak at half the task
    cost~\cite{anthropic2026opus5}.
\end{itemize}

\paragraph{Routing recovers the optimum.} If no model wins everywhere, a router
should beat every individual model. On the fourteen fully-covered benchmarks,
the best single model (GPT-5.6 Sol) averages 97.6\% of the per-benchmark best;
an oracle router averages 100\% (+2.4 points). A router with only
\emph{two} models --- GPT-5.6 Sol and Claude Fable~5 --- already captures the
entire oracle gain on this suite because their strengths are complementary
(agentic vs.\ repository-level and mathematical work). A
cost-constrained router that picks the cheapest model within one point of the
best matches the oracle score at 95\% of its cost --- the savings are modest
precisely because Sol's wins are usually by wide margins
(Section~\ref{sec:improvements} discusses implications).

\paragraph{Why models diverge.} Specialization is manufactured, not
accidental. Moonshot optimized K3 for agentic coding with a 2.5$\times$
scaling-efficiency gain from architectural changes (Kimi Delta Attention,
Attention Residuals) and a training pipeline openly aimed at the agentic
benchmark suite~\cite{moonshot2026k3}; its 1M-token single-agent setup was
deliberately tuned for long-horizon tasks like
BrowseComp~\cite{moonshot2026k3}. OpenAI trained GPT-5.6 for terminal and
multi-agent workflows (Ultra mode natively fans out
subagents)~\cite{openai2026gpt56}. And the cautionary case: Opus~5's record
ARC-AGI-3 gain partially failed to transfer to Witness, an independent
held-out puzzle benchmark (43.4, statistically tied with K3 and Fable~5),
consistent with training on genre-adjacent data~\cite{decoder2026opus5,techtimes2026opus5}.
Targeting a task family works, although gains measured on its benchmark
can overstate what transfers to new settings. Section~\ref{sec:discussion}
examines this tension.

\section{Improving fixed models at inference time}\label{sec:improvements}

We evaluate three ways to raise effective performance without retraining:
sample more than once, match reasoning effort to the workload, and route work
to specialized models.

\subsection{Sample, then aggregate}\label{sec:selfcons}

\paragraph{Why multiple attempts can help.} On problems with verifiable
answers, independent samples can expose solution paths that greedy decoding
misses. Self-consistency aggregates those attempts without changing model
weights~\cite{wang2022selfconsistency}. The selector still matters: a poor
rule can discard the only correct candidate.

\paragraph{Development and model selection.} We use the original 16-item
Qwen2.5-0.5B-Instruct run as development evidence, not a final estimate, and
lock a disjoint 100-item evaluation set before tuning. The first eight
development items screen three official-chat prompts (concise, four-shot, and
solve-and-check) at temperatures 0.4 and 0.7. We confirm the leading
four-shot/0.4 prompt on all 16 items, compare 0.5B and 1.5B models, and measure
plurality prefixes alongside a candidate verifier. A frozen manifest records
the selected configuration before evaluation begins.

\paragraph{Development results.} The legacy raw-prompt 0.5B pilot produced
1/16 greedy and 2/16 vote-at-eight. The official instruction template, four
worked examples, and temperature 0.4 raised the same model to 5/16 greedy and
6/16 with vote-at-four or the verifier. With that setup held fixed, the 1.5B
model reached 8/16 under greedy, vote-at-four, and verifier decoding on CPU.
Prompt formatting supplied the largest development gain. Model size improved
greedy and the selected vote, whereas the verifier added no development
accuracy. Because these results guided selection, we do not treat them as
unbiased performance estimates.

\begin{table}[t]
\centering
\caption{Development path and frozen evaluation counts. The legacy vote uses
$k=8$; all v1 vote rows use $k=4$. Development rows guided configuration
selection and must not be read as holdout estimates.}\label{tab:v1dev}
\small
\resizebox{\linewidth}{!}{\begin{tabular}{lrrrrr}
\toprule
Configuration & $n$ & Greedy & Vote & Verifier & Oracle \\
\midrule
Legacy raw prompt, 0.5B (dev.) & 16 & 1 & 2 & -- & -- \\
Chat + few-shot, 0.5B (dev.) & 16 & 5 & 6 & 6 & 10 \\
Chat + few-shot, 1.5B (dev.) & 16 & 8 & 8 & 8 & 10 \\
Frozen 1.5B configuration (eval.) & 100 & 58 & 62 & 62 & 79 \\
\bottomrule
\end{tabular}
}
\end{table}

\paragraph{Frozen evaluation.} The manifest fixes
Qwen2.5-1.5B-Instruct, four-shot official chat formatting, $k=4$,
temperature 0.4, top-$p=0.95$, seed 1234, and plurality as the primary
selector; it names the verifier as secondary. We run the 100 locked items once
on CUDA and retain every paired trace.
On the untouched 100-item evaluation split, greedy decoding solved 58/100 (58.0\%; Wilson 95\% CI [48.2, 67.2]), while the prespecified four-sample plurality solved 62/100 (62.0\%; Wilson 95\% CI [52.2, 70.9]). The verifier also solved 62/100 (62.0\%). The primary paired exact McNemar test gave $p=0.481$.

This four-point gain equals a 6.9\% relative improvement in the saved run, but
the confidence intervals overlap and the paired test does not reject equality.
We therefore describe an observed improvement, not an established population
effect.

\paragraph{Selection remains the bottleneck.} All 400 sampled answers were
parseable, and at least one sample was correct on 79/100 items. Plurality and
the verifier each recovered 62/100, leaving a 17-point gap to the oracle
ceiling. The verifier tied plurality on evaluation after trailing it in the
CUDA development check (6/16 versus 10/16). Complexity alone did not help; a
better selector must find correct minority candidates without disturbing
correct consensus.

\begin{figure}[t]
\centering
\submissiongraphic[width=0.92\linewidth]{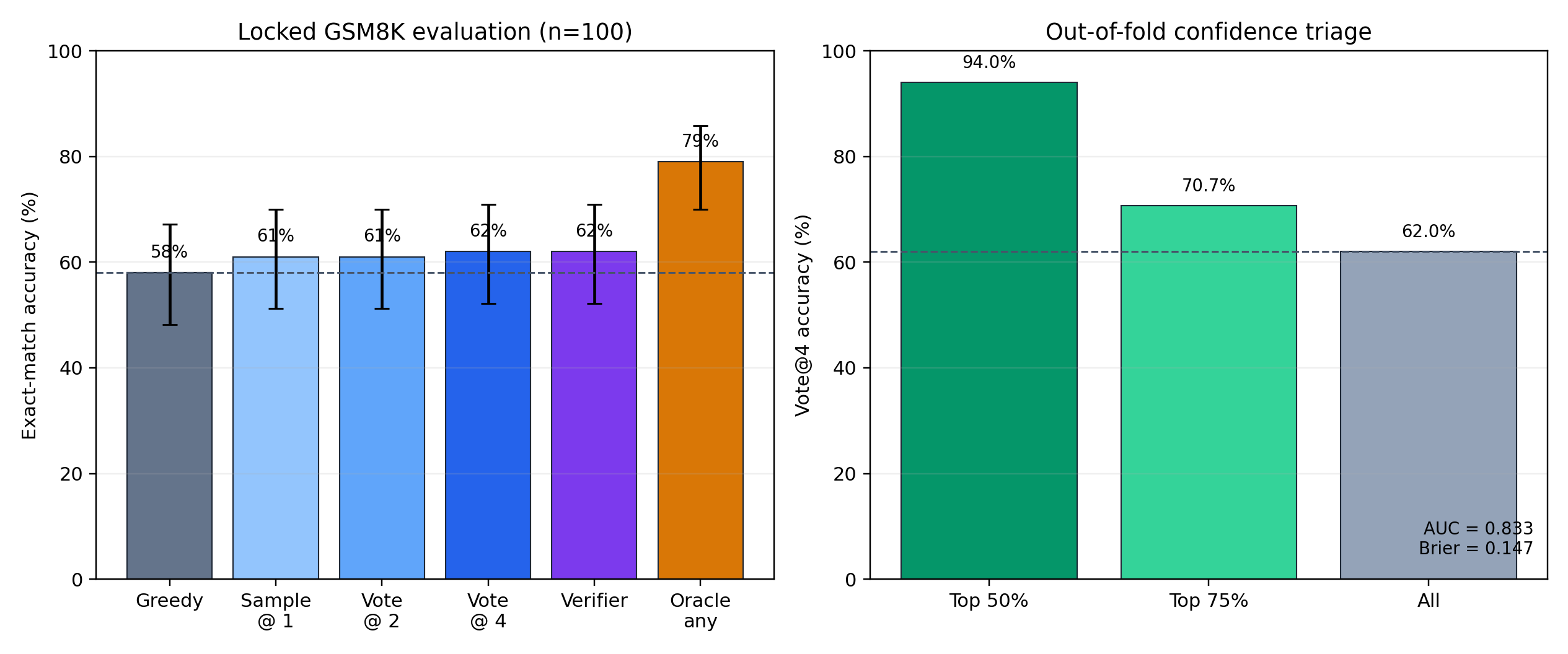}
\caption{Frozen evaluation and exploratory confidence triage. The left
panel reports exact-match accuracy with Wilson 95\% intervals; vote-at-four is
the primary result and the any-sample oracle is a diagnostic ceiling. The
right panel ranks vote-at-four predictions by out-of-fold confidence and shows
accuracy at fixed coverage. Dashed lines mark the full-set baselines.}\label{fig:selfcons}
\end{figure}

\begin{table}[t]
\centering
\caption{Frozen v1 evaluation results. Statistical inference is paired against
greedy; the primary exact McNemar test gives $p=0.481$.}\label{tab:selfcons}
\small
\resizebox{\linewidth}{!}{\begin{tabular}{lrrcl}
\toprule
Method & Correct & Accuracy & Wilson 95\% CI & Role \\
\midrule
Greedy & 58/100 & 58.0\% & [48.2, 67.2] & baseline \\
Sample @ 1 & 61/100 & 61.0\% & [51.2, 70.0] & diagnostic \\
Vote @ 2 & 61/100 & 61.0\% & [51.2, 70.0] & diagnostic \\
Vote @ 4 & 62/100 & 62.0\% & [52.2, 70.9] & prespecified primary \\
Verifier & 62/100 & 62.0\% & [52.2, 70.9] & prespecified secondary \\
Oracle: any sample & 79/100 & 79.0\% & [70.0, 85.8] & ceiling diagnostic \\
\bottomrule
\end{tabular}
}
\end{table}

\paragraph{Reproducibility.} The same seed produced different candidate
draws on CPU and CUDA, although greedy correctness remained 8/16 in the
development check. The result JSON therefore records the execution device, and
the paper uses the CUDA artifact from the 100-item evaluation. This experiment
mirrors frontier multi-sample and multi-agent inference at small scale; for
example, Sol Ultra reports 91.9\% on Terminal-Bench versus 88.8\% for the
single-agent setting~\cite{openai2026gpt56}.

\subsection{Estimate confidence for selective use}\label{sec:confidence}

After the v1 labels were available, we asked whether the traces could identify
reliable vote-at-four outputs without looking at reference answers. We fixed
seven inexpensive signals before scoring the new analysis: plurality share,
answer entropy and diversity, agreement with greedy and verifier outputs, and
two completion-length statistics. An L2-regularized logistic regression
produces one out-of-fold confidence score for each item under five-fold
stratified cross-validation.

The model separates correct from incorrect vote-at-four outputs with ROC AUC
0.833. Its Brier score is 0.147, compared with 0.236 for a constant prediction
equal to the 62\% prevalence. Confidence ranking is more useful operationally:
the top 50 predictions contain 47 correct answers (94.0\%), and the top 75
contain 53 (70.7\%). Full coverage necessarily returns to the frozen 62/100
result.

\begin{table}[H]
\centering
\caption{Exploratory selective accuracy from five-fold out-of-fold confidence
scores. Coverage levels were fixed before scoring. Top refers to confidence
rank, not a second evaluation split.}\label{tab:mlconfidence}
\small
\begin{tabular}{lrrr}
\toprule
Confidence-ranked set & Coverage & Correct & Accuracy \\
\midrule
Top half & 50/100 & 47/50 & 94.0\% \\
Top three quarters & 75/100 & 53/75 & 70.7\% \\
All items & 100/100 & 62/100 & 62.0\% \\
\bottomrule
\end{tabular}

\end{table}

A deployed system could accept the high-confidence half and send the remainder
to additional sampling, a larger model, or human review. This procedure
improves the accuracy of the retained set, not full-coverage exact match.
Because the same 100 labels both train and assess the cross-validated model,
independent data must confirm the result.

\subsection{Match reasoning effort to the workload}\label{sec:effort}

Maximum effort can cost more while producing a worse answer. Anthropic's
score--cost curves put Opus~5's Frontier-Bench peak at \emph{xhigh} effort
(44.4\%); max effort falls to 43.3\% despite consuming more compute. A
community re-analysis of Anthropic's Frontier-code chart places that peak near
medium effort (${\sim}53\%$)~\cite{anthropic2026opus5}. Our Pareto analysis
shows a different part of the same tradeoff: each roughly four-point
intelligence gain from Luna-max to Terra-max to Sol-max multiplies per-task
cost by $2.6\times$ and then $1.9\times$~\cite{gist2026aa56}. Longer chains
can wander on tasks with direct solution paths. Deployments should therefore
measure effort per workload and use its empirical optimum rather than treating
maximum effort as a universal quality setting.

\subsection{Route work across specialized models}\label{sec:routing}

The fragmented frontier in Section~\ref{sec:specialization} makes routing more
effective than a single default model. Across the fourteen fully covered
benchmarks, the best individual model averages 97.6\% of the per-benchmark
maximum. An oracle reaches 100, and a two-model combination of Sol and Fable~5
captures that entire gain. Production systems increasingly expose this choice
through model-abstraction layers~\cite{explainx2026gpt56}; practitioners also
pair a strong planner with a cheaper executor, such as ``Sol medium for
architecting, Luna high for coding''~\cite{hn2026gpt56}. Two complementary
frontier models can cover quality-sensitive work, while a budget tier handles
well-defined subtasks at lower cost.

\subsection{Reporting priorities for model builders}\label{sec:labs}
Model builders can make these tradeoffs easier to evaluate by publishing
score--cost curves for every benchmark; single endpoint numbers invite
cherry-picking~\cite{anthropic2026opus5}. Budget models also need better
long-context recall: Luna scores 41.3\% on MRCR against Sol's
91.5\%~\cite{openai2026gpt56}, the clearest break from its near-flagship
pattern. Held-out transfer suites should accompany targeted benchmarks, as the
ARC-AGI-3/Witness gap in Section~\ref{sec:specialization} demonstrates.
K3's BrowseComp result further points toward better mid-context retrieval, not
longer context windows alone, as the more valuable investment
target~\cite{moonshot2026k3}.

\section{Discussion: how models get good at specific tasks}\label{sec:discussion}

The 2026 pattern --- one frontier, several winners --- follows directly
from the way laboratories now build frontier models.

\paragraph{Domain-weighted post-training.} Reinforcement learning on
verifiable, domain-specific rewards (unit tests for code, proof checkers for
math, task-completion signals for agents) lets a lab buy capability in a chosen
vertical. Moonshot's agentic suite results and OpenAI's terminal/agentic lead
both track heavy RL investment in those task
families~\cite{moonshot2026k3,openai2026gpt56}.

\paragraph{Architecture as targeting.} Kimi~K3's hybrid linear attention
(Kimi Delta Attention) and expert-routing changes were chosen for long-context
agentic efficiency, and the payoff appeared precisely on long-horizon
benchmarks (BrowseComp 91.2, state of the art at
release)~\cite{moonshot2026k3}.

\paragraph{Inference-time structure.} GPT-5.6's Ultra mode makes
multi-agent fan-out a first-class feature and buys Terminal-Bench points
(88.8\%~$\to$~91.9\%) at higher cost~\cite{openai2026gpt56}. Our
self-consistency experiment (Section~\ref{sec:selfcons}) applies the same lever
at small scale. The exploratory confidence model adds a routing signal: accept
high-confidence answers, then reserve additional compute or review for the
rest. Capability depends partly on how a system orchestrates inference.

\paragraph{Benchmark-shaped optimization and its limits.} When a benchmark
becomes a target, labs optimize toward it, and two failure modes follow: the
benchmark saturates (GLUE, MMLU, soon SWE-bench Verified), or gains fail to
transfer off-distribution. The Witness evaluation of Opus~5 is the cleanest
2026 example: a ${\sim}4\times$ record on ARC-AGI-3, but only a statistical tie
with K3 and Fable~5 on held-out puzzles of the same genre, including one
environment where Opus~5 stated the hidden rules before its first move ---
suggesting familiarity, not just reasoning~\cite{decoder2026opus5,techtimes2026opus5}.
METR's maintainer-review study adds a second caution from a different angle:
a growing share of SWE-bench-passing patches would be rejected by human
maintainers (quality, breakage, or missing the issue's point), and the
grader--maintainer gap is widening over model
generations~\cite{metr2026swebench}. Targeting works, and that is exactly why
single-number claims deserve suspicion.

\paragraph{Implication for evaluation.} Score-versus-cost curves, held-out
transfer suites, and maintainer-grade review are becoming the minimum standard
of evidence. Vendors already report curves~\cite{anthropic2026opus5}; ARC Prize
now administers runs independently with published
traces~\cite{arcprize2026opus5, techtimes2026opus5}; and METR's maintainer
study is a template for auditing what benchmark passes actually mean.

\paragraph{Implication for buyers.} The economic shape of 2026 --- a budget
tier at 20\% of flagship token price that matches last quarter's flagship on most
professional work, plus a fragmented frontier --- means the unit of procurement
is no longer ``a model'' but a routing policy with per-task effort settings.
The 2023 question ``which model is best?'' has been replaced by ``which
combination, at which effort, for which task?''

\paragraph{Limitations.} Mid-2026 scores mix vendor and independent
evaluations. Provenance labels cannot fully remove harness differences:
GPT-5.5, for example, scores 85.1\% in the vendor report and 82.6\% in our
reconstruction of Vals AI's independent
harness~\cite{localaimaster2026gpt55,vals2026swebench}. The SWE-bench trend
also pools both sources across only 22 months, so $5.8\times$ indicates scale
rather than a stable law. Our routing oracle knows benchmark identity; a
production router must infer task type from unlabeled traffic.

The v1 intervention used 16 development items and one 100-item GSM8K
evaluation. It covers a single model family, task, and sampling seed. Its
four-point primary gain is not statistically significant ($p=0.481$), and the
Wilson intervals remain broad. We designed the confidence model after those
labels were visible and assessed it by cross-validation on the same 100 items.
Its AUC and 94\% top-half accuracy are exploratory until a new dataset confirms
them. Seeded candidates also differ between CPU and CUDA, so exact replay
requires the recorded software and device context. Public benchmark figures
are current as of August 10, 2026, in a field that changes weekly.

\section{Conclusion}\label{sec:conclusion}

Eight years separate BERT's 80.5 GLUE score from Opus~5's reported 42/42 on
the 2026 IMO problems. Progress did not follow one smooth exponential. Each
shift --- fine-tuning, few-shot scale, alignment, reasoning, and agency ---
moved capability onto a new axis and exhausted the previous era's benchmarks.
Agentic-coding solve odds are still rising at roughly six-fold per year, while
the price of a fixed capability level falls fast enough that a \$1/M budget
tier now reproduces most of the professionally relevant performance of an
eleven-week-old flagship. Different models lead frontend coding, repository
repair, terminal work, and novel reasoning, so deployment increasingly depends
on routing policy rather than one model choice.

Fixed weights still leave room for system-level gains. In our frozen GSM8K
evaluation, four-sample plurality moved exact match from 58\% to 62\%, though
the four-point difference remains uncertain ($p=0.481$). At least one sample
was correct on 79\% of items, which makes selection the clearest next target.
An exploratory logistic model found a useful signal in answer agreement and
trace statistics: out-of-fold AUC reached 0.833, and 47 of its 50
highest-confidence predictions were correct. That result supports selective
triage, not a claim that full-coverage accuracy exceeds 62\%, and it still
requires confirmation on untouched data.

\paragraph{Reproducibility.} The accompanying archive contains all
datasets (\texttt{code/data/}), analysis scripts
(\texttt{exp1}--\texttt{exp5}), locked manifests, complete inference traces,
unit tests (\texttt{tests/}), figure generation, and the paper build script;
see \texttt{README.md}. Reproduction needs no proprietary access beyond the
live benchmark numbers, whose sources and retrieval dates appear in
\texttt{code/data\_sources.md}.

\begingroup
\emergencystretch=2em
\bibliography{references}
\endgroup

\end{document}